\documentclass{article} 
\usepackage{iclr2025_conference,times}

\usepackage{amsmath,amsfonts,bm}

\def\eqref#1{equation~\ref{#1}}

\def\1{\bm{1}}

\DeclareMathAlphabet{\mathsfit}{\encodingdefault}{\sfdefault}{m}{sl}
\SetMathAlphabet{\mathsfit}{bold}{\encodingdefault}{\sfdefault}{bx}{n}

\usepackage{hyperref}
\usepackage{url}
\usepackage{caption}
\usepackage{graphicx}
\usepackage{wrapfig}
\usepackage{multirow} 
\usepackage{multicol}
\usepackage{booktabs}
\usepackage{subcaption}
\usepackage{authblk}
\usepackage[skip=1pt]{caption}

\title{Hierarchical Multi-Stage Transformer Architecture for Context-Aware Temporal Action Localization}

\author[1]{{\textbf{Hayat Ullah}}}
\author[1]{{\textbf{Arslan Munir}}}
\author[2]{{\textbf{Oliver Nina}}}
\affil[1]{Department of Electrical Engineering and Computer Science, Florida Atlantic University}
\affil[2]{Air Force Research Laboratory}
\affil[1]{\texttt{\{hullah2024,arslanm@\}@fau.edu}}
\affil[2]{\texttt {oliver.nina.1@afrl.af.mil}}

\iclrfinalcopy 
\begin{document}

\maketitle

\begin{abstract}
Inspired by the recent success of transformers and multi-stage architectures in video recognition and object detection domains. We thoroughly explore the rich spatio-temporal properties of transformers within a multi-stage architecture paradigm for the temporal action localization (TAL) task. This exploration led to the development of a hierarchical multi-stage transformer architecture called PCL-Former, where each subtask is handled by a dedicated transformer module with a specialized loss function. Specifically, the Proposal-Former identifies candidate segments in an untrimmed video that may contain actions, the Classification-Former classifies the action categories within those segments, and the Localization-Former precisely predicts the temporal boundaries (i.e., start and end) of the action instances. To evaluate the performance of our method, we have conducted extensive experiments on three challenging benchmark datasets: THUMOS-14, ActivityNet-1.3, and HACS Segments. We also conducted detailed ablation experiments to assess the impact of each individual module of our PCL-Former. The obtained quantitative results validate the effectiveness of the proposed PCL-Former, outperforming state-of-the-art TAL approaches by 2.8\%, 1.2\%, and 4.8\% on THUMOS14, ActivityNet-1.3, and HACS datasets, respectively.
\end{abstract}

\section{Introduction}
\label{sec:intro}
Temporal action localization (TAL) plays a key role in understanding the context of long videos containing both action-specific and background scenes. TAL in untrimmed videos involves identifying and localizing actions within a video timeline without prior segmentation or labeling. It's a critical task in video understanding to precisely determine actions' occurrence and their duration within long untrimmed video sequences. However, temporal localization of action-specific scenes in untrimmed videos is a challenging problem due to various complexities. One of the primary challenges lies in the inherent variability of actions within untrimmed videos. Actions can occur in different temporal extents, speeds, and occlusions, making their accurate detection and localization a formidable task. This variability necessitates models capable of handling these nuances while discerning actions from complex backgrounds, diverse contexts, and occluded instances, thereby demanding robustness and adaptability from TAL algorithms.\\ \\
\indent \indent To address these challenges, various TAL approaches have been developed, broadly categorized into three types: (1) \textit{Anchor-based}, (2) \textit{Actionness-guided}, and (3) \textit{Anchor-free} approaches. Anchor-based methods~\cite{chao2018rethinking, gao2017turn, xu2017r} generate action proposals using predefined anchors across temporal scales, followed by classification and boundary regression. Contrarily, Actionness-guided methods~\cite{lin2018bsn, lin2019bmn} predict action-start, action-end probabilities, and duration to form proposals without predefined anchors. Anchor-free methods~\cite{lin2021learning, ou2022srfnet} generate proposals at each temporal position by associating the start and end of action, reducing proposal redundancy and computational overhead compared to action-ness-guided approaches.\\ \\
\indent We observe several notable limitations in existing TAL approaches. First, anchor-based methods~\cite{chao2018rethinking, gao2017turn, xu2017r} rely on predefined anchors, limiting their adaptability to varying action lengths and making them sensitive to hyper-parameters. Second, actionness-guided methods ~\cite{lin2018bsn, lin2019bmn} remove this dependency but introduce significant computational overhead due to the need for an additional classification model. Third, Anchor-free approaches~\cite{lin2021learning, ou2022srfnet} reduce redundancy and computational cost yet may struggle with long-range temporal dependencies and handling actions of varying durations. Overall, these methods face challenges in balancing accuracy, efficiency, and adaptability across different action categories and temporal scales.\\ \\
\indent Inspired by the recent success of multi-stage transformer architectures in the object detection domain \cite{zhang2023towards}, we thoroughly explore transformer for temporal action localization task and propose three-tier transformer architecture named PCL-Former. Unlike existing TAL methods, which utilize pre-trained video encoders, these models fail to fully leverage the inherent semantic information due to limited feature representation. Specifically, video encoders pre-trained for video-level classification are not optimized for the TAL task, resulting in snippet-level features that lack sufficient contextual information for precise action localization. Our PCL-Former is designed to address three core subtasks of TAL using distinct transformer modules. The Proposal-Former identifies candidate segments that may potentially contain actions and eliminates background segments. The Classification-Former classifies the action category in a given segment. It is worth mentioning here that the classification network focuses on categorizing action instances rather than accurately localizing their temporal boundaries. In some cases, high classification scores may be assigned to segments with minimal overlap with the ground truth, which can be problematic. Post-processing techniques like Non-Maximum Suppression (NMS) may discard segments with small scores but significant overlap with the ground truth. To address this, we introduce the Localization-Former, equipped with a specialized loss function that prioritizes segments with higher temporal overlap. This encourages more reliable confidence scores for downstream post-processing, improving action localization accuracy.\\ \\
\indent More precisely, the main \textbf{contributions} of this paper are the following: (1) To overcome the fundamental limitations of mainstream TAL methods, such as limited adaptability to varying action lengths, high computational overhead of independently trainable modules, and challenges with long-range temporal dependencies, we leverage the rich spatio-temporal modeling capabilities of transformers and multi-stage architectures for the temporal action localization task. (2) We propose PCL-Former, an effective three-tier transformer architecture designed to handle the key subtasks of TAL with specialized transformer modules, each guided by a task-specific loss function. Specifically, the Proposal-Former generates candidate segments likely to contain actions, the Classification-Former categorizes actions in a given segment, and the Localization-Former predicts the temporal boundaries of actions within the segment. During testing, the Proposal-Former effectively filters out background segments, enhancing overall performance. (3) The proposed method is extensively evaluated on three benchmark datasets commonly used for the temporal action localization task. The obtained quantitative results validate the effectiveness of the proposed PCL-Former, outperforming state-of-the-art TAL approaches by 2.8\%, 1.2\%, and 4.8\% on THUMOS14, ActivityNet-1.3, and HACS datasets, respectively.

\section{Related Work}
\label{sec:related_work}
\indent \indent \textbf{Action Recognition:} In the realm of video comprehension, action recognition stands as a fundamental task, significantly advanced by deep learning methods. Leveraging 2D and 3D CNNs \cite{tran2015learning, carreira2017quo, wang2016temporal, tran2018closer} has notably boosted performance on standard action recognition benchmarks. More recently, Vision Transformer-based approaches \cite{bertasius2021space, liu2022video, fan2021multiscale} have shown considerable superiority over previous CNN-based methods. These Vision Transformer models, benefiting from extensive pre-training on action recognition datasets, are commonly employed as short-term feature extractors in temporal action localization tasks. However, a drawback of these contemporary video transformer models is their slower training pace and hefty GPU memory requirements due to the quadratic memory complexity of self-attention mechanisms \cite{vaswani2017attention}. Consequently, their effective application to long-term modeling tasks, such as temporal action localization, poses notable challenges.\\ \\
\indent \textbf{Temporal Action Localization:}
The temporal action localization approaches \cite{zhao2022tuber,faure2023holistic,lin2017single, long2019gaussian,xu2017r,chao2018rethinking} fall into two distinct classes that include \textit{Single-Stage TAL} and \textit{Two-Stage TAL} approaches. The \textit{Single-Stage} TAL methods localize actions in a video without any action proposal, generating anchors using a sliding window. The SSAD \cite{lin2017single} and GTAN \cite{long2019gaussian} are Single-Stage methods. The SSAD \cite{lin2017single} employed a single-shot structure using 1D convolution to generate anchors for temporal action localization, while GTAN \cite{long2019gaussian} utilized a 3D ConvNet to extract fine-grained segment-level features. GTAN further employs temporal Gaussian kernels to generate proposals with varying temporal resolutions. On the other hand, Two-Stage TAL approaches \cite{xu2017r, chao2018rethinking} first generate candidate segments from input video as action proposals and then classify them into actions and refine their temporal boundaries. The R-C3D \cite{xu2017r} method integrates candidate segment generation and classification within an end-to-end network, allowing input videos of varying lengths. TALNet \cite{chao2018rethinking} specifically focused on expanding the receptive field for feature extraction and incorporates temporal context extraction alongside late fusion in a two-stream architecture. \\ \\
\indent \textbf{Transformers in Videos:}
The transformer architecture, originally introduced for the Natural Language Processing (NLP) task in \cite{vaswani2017attention} reliant on self-attention mechanisms to capture sequence relationships, has recently gained substantial interest in computer vision. Specifically, the self-attention mechanism within transformers has emerged as a pivotal tool for extracting temporal information from videos. In object detection, DETR \cite{carion2020end} innovatively employed object queries instead of traditional anchors as candidates. Yet, for action detection, a pure transformer architecture might prove insufficient in handling intricate temporal dependencies. MS-TCT \cite{dai2022ms}, recognizing this limitation, incorporates convolutions to enable multiple temporal scales of tokens, ensuring temporal consistency and integrating local information between tokens. In the domain of Temporal Action Localization, ATAG \cite{chang2022augmented} introduces an augmented transformer with an adaptive graph network, capturing both long-term and local temporal context for enhanced understanding. Similarly, TAPG \cite{wang2021temporal} devises a unified framework comprising boundary and proposal transformers, each focusing on long-term dependencies and proposal relationship enrichment, ensuring accurate boundary prediction and reliable confidence assessment. Inspiring from ViT \cite{dosovitskiy2020image}, ViViT \cite{arnab2021vivit} introduced a factorized encoder model a fusion of spatial and temporal transformer encoders to model diverse temporal interactions, enhancing spatial-temporal understanding. \\ \\

\section{Our Method}
\label{sec:method}
\begin{figure}[t]
  \centering
  \includegraphics[width=1.0\linewidth]{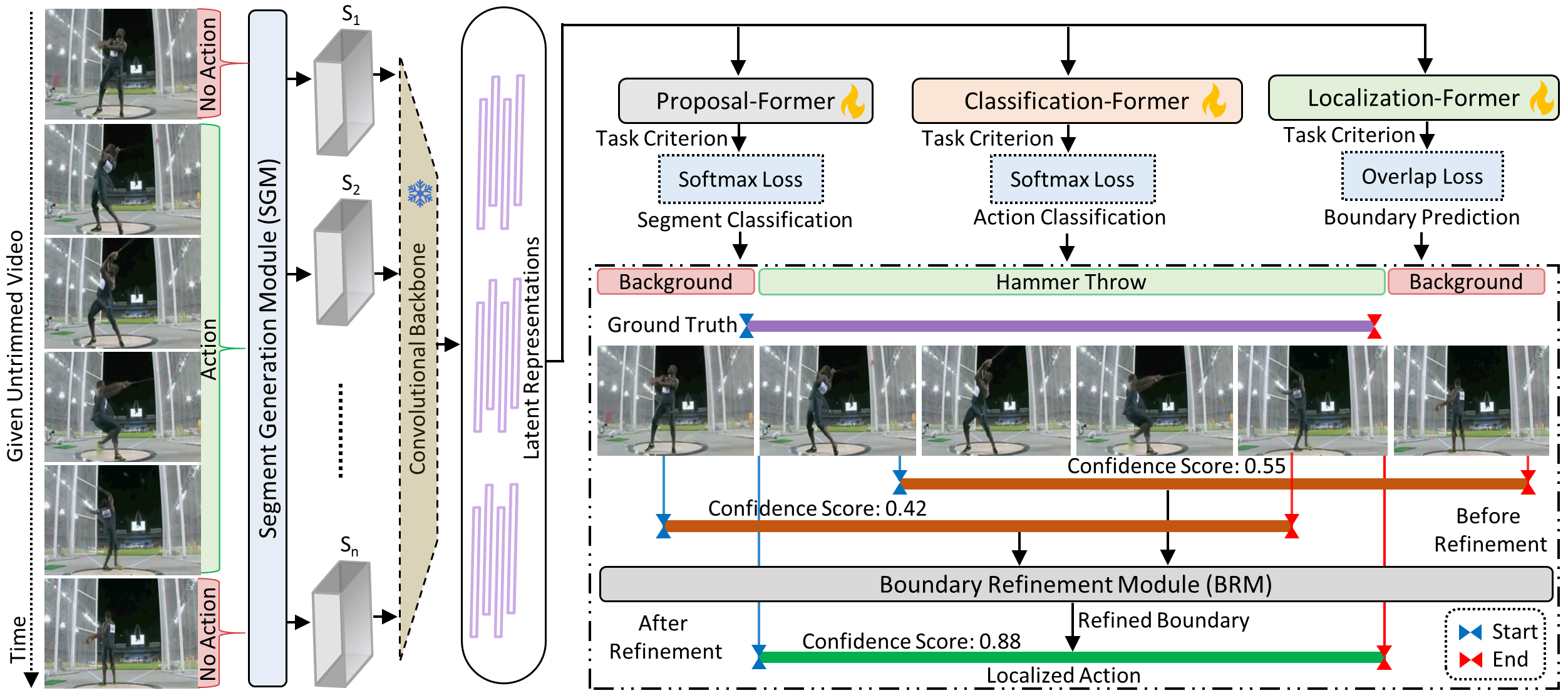}
  \caption{Visual overview of our proposed PCL-Former. The input untrimmed video is divided into fixed-size segments by the SGM module. These segments are passed to a convolutional backbone for video-level feature extraction. The latent representations are then fed into sub-task transformers. The outputs (action class and boundaries) are refined by the BRM module to produce the final predictions.}
  \label{fig:framework}
\end{figure}

\indent \indent In this section, we provide the problem formulation and a detailed explanation of our proposed PCL-Former architecture. The detailed visual overview of our proposed method is depicted in Figure \ref{fig:framework}.
\subsection{Problem Formulation}
\indent \indent The temporal action localization task involves two sub-tasks: action category classification and action interval (start and end timestamps of action instance) prediction in a given untrimmed video. To better formulate this problem, consider a given untrimmed video $\mathbf{X} \in \mathbb{R}^{3\times H \times W \times T}$, where $H$ and $W$ are the height and width, and $T$ is the number of frame or temporal length. The temporal actions can be denoted as ${{\Psi }_{g}}=\left\{\varphi _i\!=\!(t_s,t_e,c)\right\}_{i=1}^N$, where $t_{s}$, $t_{e}$, $c$ are the start, end time and category of action instance $\varphi _i$, and $N$ is the number of actions in a given untrimmed video. Further, the action instance $\varphi _i$ starts at time $\varphi_{i}^{t_{s}}$ that is $(0\leq \varphi_{i}^{t_{s}} <N)$ and ends at time $\varphi_{i}^{t_{e}}$ that is $(0 < \varphi_{i}^{t_{e}} \leq N)$. Thus, our objective is to predict the category $c$ of action instance $\varphi_{i}$ and its occurrence interval ($t_{s}$ and $t_{e}$) in a given untrimmed video.
\subsection{PCL-Former}
\indent \indent In pursuit of this objective, we propose a three-tier transformer architecture called PCL-Former for solving three distinct tasks inspired by \cite{shou2016temporal}. \textit{Proposal-Former} for generating candidate segments, \textit{Classification-Former} for action instance classification, and \textit{Localization-Former} for action boundary prediction. Besides, our PCL-Former includes a pre-processing module called \textit{Segment Generation Module} (SGM) that segments the input untrimmed video in multiple segments. The convolutional backbone extracts video-level features from input segments. A post-processing module called \textit{Boundary Refinement Module} (BRM) for refining the predicted boundaries for precise localization of action instances. \\ \\
\indent It is worth mentioning here that each sub-network of PCL-Former utilizes the ViViT-like architecture \cite{arnab2021vivit} due to its overwhelming performance on video classification tasks. However, instead of standard multi-head self-attention, we use factorized self-attention as it reduces the quadratic complexity of computing attention scores to linear complexity, making it more scalable for long sequences. This is achieved by decomposing the attention computation into smaller, more manageable steps, which decreases computational and memory overhead while maintaining the overall model's performance. \\ \\
\indent \textbf{Segment Generation Module:} Given an untrimmed video $\mathcal{V}= \Sigma_{i=1} ^ \mathcal{N} (f_{i})$, we divide it into fixed size (64 frames per segment) multiple segments $\mathcal{S} = \{s_{1},s_{2},...s_{n}\}$, by traversing temporal sliding window over the length of $\mathcal{V}$, having 50\% overlap between two consecutive segments, where the frame resolution; height $\mathcal{H}$ and width $\mathcal{W}$ are set to 224 and 224, respectively. Thus for each untrimmed video $\mathcal{V}$, we get a list of candidate segments $\mathcal{S}$, where each segment has a length of 64 frames having a resolution of ($224 \times 224$) with 50\% frames from the previous segment.
\begin{equation}
\mathcal{S}\{s_{1},s_{2},...s_{n}\} = \sum_{i=1}^{N} SGM(\mathcal{V})_{r} ^ {t}
\label{Eq:equation_1}
\end{equation}
Where $t$ is the temporal length of each segment (i.e., 64) and $r$ is the resolution (i.e., $224 \times 224$) of frames in each segment. \\ \\
\indent \textbf{Proposal-Former:} The main objective of \textit{Proposal-Former} $\Theta_{PF}$ in this study is to filter the input segment for background and action instances. To prepare the training data for $\Theta_{PF}$, we follow the data preparation protocol given in \cite{shou2016temporal}. Let, $D_{prop}^{train}$ = \{($s_{i}$,$c_{i}$)\}, where $c_{i}$ $\in$ \{0,1\} is the action category of $i^{th}$ instance. 
\begin{figure}[t]
    \centering
    \includegraphics[width=\linewidth]{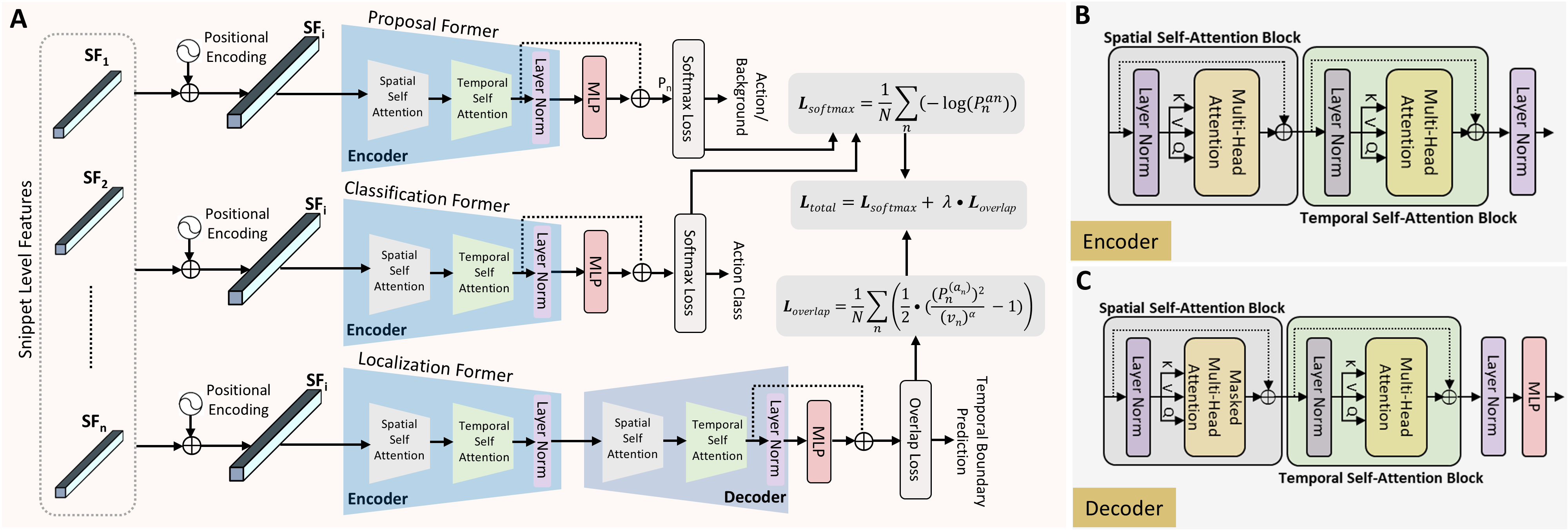}
    \caption{The detailed architectural overview of our \textbf{PCL-Former}: \textbf{(A}) Depicts the complete workflow of training protocol of our entire architecture, processing the input features through \textbf{Proposal Former}, \textbf{Classification Former}, and \textbf{Localization Former} equipped with task-specific loss functions. Moreover, (\textbf{B}) depicts the block diagram of \textbf{Encoder} module used in each sub-network and (\textbf{C}) depicts the block diagram of \textbf{Decoder} module used in Localization Former.}
    \label{fig:architecture}
\end{figure}
\begin{wrapfigure}{r}{8.5cm}
    \centering
    \includegraphics[width=0.60\textwidth]{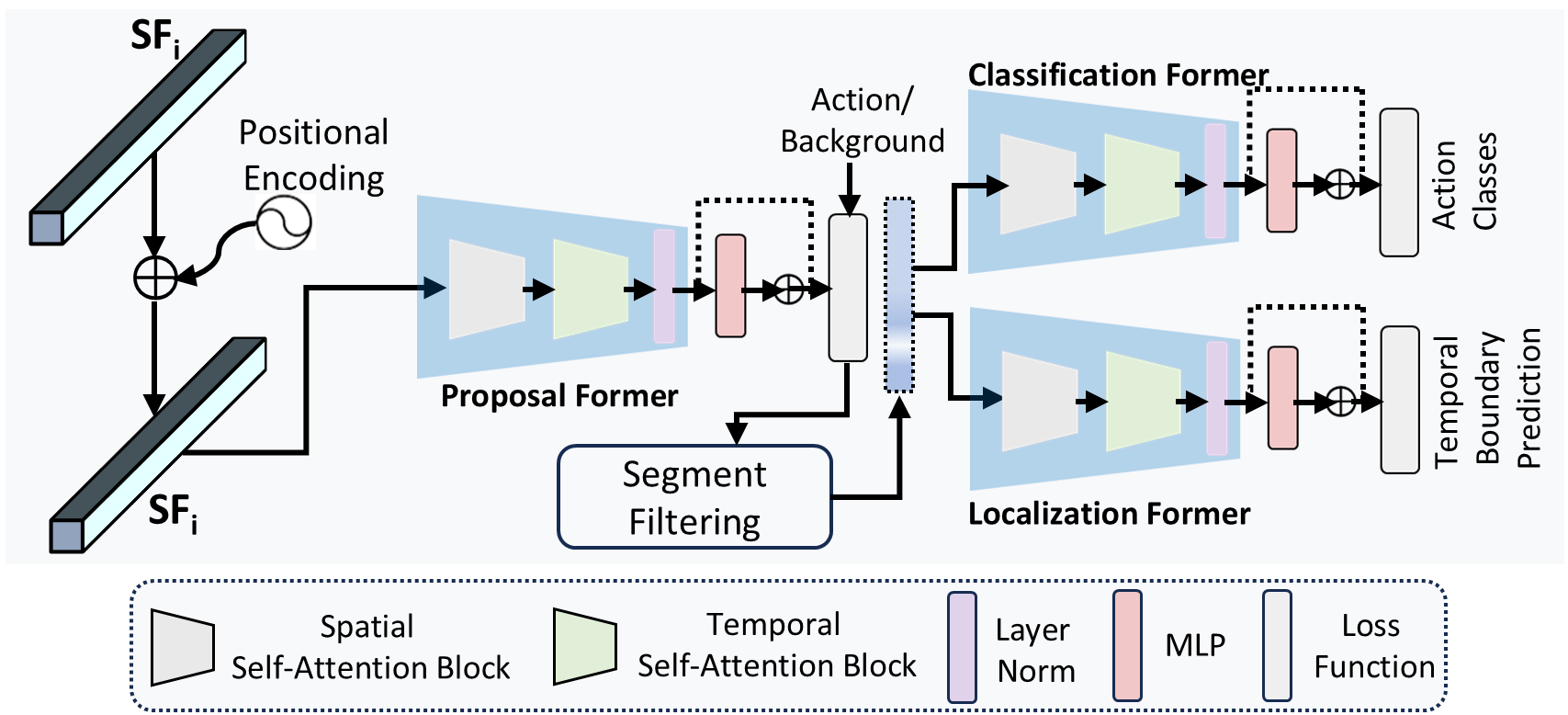}
    \caption{The detailed inference workflow of our \textbf{PCL-Former} architecture: the \textbf{Proposal Former} identifies candidate segment features ($SF_{i}$), segment filtering discards background segments, and only positive (action-containing) segments are passed to the \textbf{Classification Former} and \textbf{Localization Former} for action classification and temporal boundary prediction.}
    \label{fig:inference}
\end{wrapfigure}
For a given set of segments $\mathcal{S}$ of an untrimmed video $\mathcal{V}$ along with its temporal annotation $\mathfrak{A}$, labels are assigned to each segment by evaluating its overlap with the ground truth using Intersection Over Union (IoU). Segments having IoU greater than 0.7 are labeled as positive and lower than 0.3 are labeled as negative. Given the input set of segments $\mathcal{S}$ from $D_{prop}^{train}$, we extract features using convolutional backbone (TSN \cite{wang2016temporal} and I3D \cite{carreira2017quo}). the $\Theta_{PF}$ performs linear positional encoding on the extracted features and passes it to the sequential multi-head self-attention blocks, where each block contains a multi-head attention layer, normalization layer, and MLP block. The output embedding of the MLP block is then passed to \textit{softmax} layer for binary classification (i.e., background ($b$) or action ($a$)). \\ \\
\textbf{Classification-Former:} The \textit{Classification-Former} $\Theta_{CF}$ has the same architecture as the \textit{Proposal-Former} except the last layer, as shown in Figure \ref{fig:architecture}. Unlike the previous task, where the goal is to differentiate between background and action scenes. Here the objective of \textit{Classification-Former} is to not only differentiate between background and action scenes but also predict the category label of action in the presence of action ($a_{c},\;c \in [1,2,....,\mathcal{C}]$). Where $\mathcal{C}$ is the number of action classes. Training data for $\Theta_{CF}$ is prepared using the same approach as for $\Theta_{PF}$. Except with one change, when labeling positive segments, each segment is labeled with an action category. \\ \\
\indent \textbf{Localization-Former:} The Localization-Former $\Theta_{LF}$ operates on the input segments and prioritizes segments with higher overlap with the ground truth instances. Elevating the prediction score $\mathcal{P}$ for segments that closely align with the ground truth and reducing scores for those with lesser overlap using overlap loss given in Eq.\ref{Eq:equation_3}, which ensures a refined selection process in the subsequent post-processing steps. The $\Theta_{LF}$ is trained on temporally annotated data $D_{Loc}^{train}$, prepared by associating each segment $s$ $\in$ $\mathcal{S}$ with its overlap $v$. If a given segment $s$ has positive action $a$ $\in$ $\mathcal{A}$, $v$ is assigned with the overlap value between the segment $s$ and its corresponding ground truth $\mathcal{GT}$. If a segment $s$ is a background segment, we set $v$ equal to 0. The encoder part of $\Theta_{LF}$ is the same as Proposal- and Classification-Former, except MLP block, which is excluded, as depicted in Figure \ref{fig:architecture}(A). On the other hand, in the decoder architecture, we replaced the spatial multi-head self-attention with multi-head masked self-attention, as shown in Figure \ref{fig:architecture}(C). The temporal multi-head self-attention is used as it is used in the encoder. Upon receiving the encoded representations from the encoder, the decoder employs masked multi-head self-attention to manage both current and preceding outputs. \\ \\
\indent \textbf{Boundary Refinement Module:} The boundary refinement module is used as a post-processing module for filtering the predicted actions and their boundaries (confidence score). The BRM module first eliminates all segments having less overlap with ground truth. Then it proceeds to enhance the remaining segments by cross-referencing them with the occurrence of actions within the corresponding time intervals. Finally, the redundant detections (temporal boundaries) are removed using Non-Maximum Suppression (NMS). Moreover, in training phase, all three sub-networks are trained with their associated task-specific loss functions (as given in Eq. (\ref{Eq:equation_2}) and Eq. (\ref{Eq:equation_3})) in unified training settings, the weighted sum of each loss is used to train the overall PCL-Former model as given in Eq. \ref{Eq:equation_4}. While testing, the Classification-Former and Localization-Former operate on the candidate segments identified by the Proposal-Former as shown in Figure~\ref{fig:inference}. \\ \\
\indent \textbf{Loss Functions:} For candidate segments (having actions not background) generation and action classification tasks, we have used softmax loss to train the \textit{Proposal-Former} and \textit{Classification-Former}. Mathematically, the softmax loss can be expressed as follows:
\begin{equation}
\mathcal{L}_{softmax} = \frac{1}{N} \sum_{n} (- \log (P_{n}^{a_{n}}))
\label{Eq:equation_2}
\end{equation}
\indent Here, $P$ is the probability vector, containing prediction voting for each class ($P_{n}$ prediction probability of $n^{th}$ class). To train the \textit{Localization-Former} for action localization task, we utilized a specialized loss function called \textit{Overlap Loss} introduced in \cite{shou2016temporal}. This function inherently incorporates Intersection over Union (IoU) with ground truth instances, enhancing the model's capacity to discern segments with higher overlaps, thereby optimizing subsequent selection processes. Mathematically, the overlap loss can be equated as follows:
\begin{equation}
\mathcal{L}_{Overlap} = \frac{1}{N} \sum_{n} (\frac{1}{2} \cdot (\frac{(P_{n}^{(a_{n})})^{2}}{(v_{n})^{\alpha}} - 1))
\label{Eq:equation_3}
\end{equation}
\indent Where $a_{n}$ is the action category of $n^{th}$ instance in a segment. In case of \textit{Proposal-Former} $a_{n = 1}$ refers to action and $a_{n = 0}$ refers to background. While in the case of \textit{Classification-Former} $a_{n}$ indicates the $n^{th}$ action category. Thus, the overlap loss serves the purpose of improving the temporal boundary prediction scores $P$ for segments that exhibit maximum overlaps with the ground truth instances while diminishing the scores for those with minimal overlaps. The hyper-parameter $\alpha$ regulates the degree of confidence score. For unified training, the aforementioned losses are combined as follows:
\begin{equation}
\mathcal{L}_{total} = \mathcal{L}_{softmax} + \lambda \cdot \mathcal{L}_{Overlap}
\label{Eq:equation_4}
\end{equation}
\indent Here, the hyper-parameter $\lambda$ ensures the balance contribution of each loss in the final loss $\mathcal{L}_{total}$.

\section{Experiments and Results}
\label{sec:results}
\indent \indent To evaluate the effectiveness of the proposed PCL-Former, we have conducted extensive experiments on three commonly used benchmark datasets. First, we briefly describe the datasets used in this work, followed by the evaluation metrics used for quantitative evaluation. Next, we briefly discuss the implementation details and experimental settings.
\subsection{Datasets}
\indent \indent We conducted experiments on three TAL benchmark datasets, which include THUMOS14 \cite{idrees2017thumos}, ActivityNet-1.3 \cite{caba2015activitynet}, and HACS \cite{zhao2019hacs} dataset. The \textbf{THUMOS14} dataset contains 413 untrimmed videos containing 20 temporally annotated actions. Following \cite{gao2022temporal}, we used 200 untrimmed videos for training and the remaining 213 for testing/evaluation. The \textbf{ActivityNet-1.4} dataset contains 20,000 untrimmed videos spanning across 200 action categories. These videos are distributed among training, validation, and testing sets in a ratio of 2:1:1, respectively. Following \cite{lin2018bsn, zhang2022actionformer}, We train our model using the training set and evaluate its performance on the validation set. The \textbf{HACS} dataset contains 140K complete action segments across 50,000 videos. Following the standard training protocol, we use the training set and evaluate the model on the validation set.
\subsection{Evaluation Metrics}
\indent \indent We used Mean Average Precision (mAP) as the performance evaluation metric for quantitative assessment and comparison with state-of-the-art approaches. To better generalize the performance of our method, we evaluated the temporal action localization task with different tIoU thresholds. For the THUMOS14 dataset, we used a set of tIoU thresholds including \{0.1, 0.2, 0.3, 0.4, 0.5, 0.6, 0.7\}. For ActivityNet-1.3 and the HACS dataset, we used tIoU thresholds including \{0.5, 0.75, 0.95\}.
\subsection{Implementation Details}
\indent \indent All experiments are conducted on a Linux-based (Ubuntu) machine having a 3.4GHz processor (40 cores), 128GB of onboard RAM, and two Tesla P4 8GB GPUs operated with CUDA 11.3. Further, the proposed method is implemented in PyTorch V1.12 using a well-known codebase MMAction2 \cite{2020mmaction2}. The proposed PCL-Former is trained for 100 epochs on the THUMOS14 dataset using TSN \cite{wang2016temporal} and I3D \cite{carreira2017quo} backbones with a learning rate 1e-4, batch size 1 (single untrimmed video), and a weight decay of 1e-4. Similarly, for ActivityNet-1.3 and the HACS dataset, we have trained our model for 100 epochs using I3D \cite{carreira2017quo} backbone with a learning rate 1e-4, batch size 1 and a weight decay of 1e-4. It is worth noting here that we have used only RGB videos from each dataset to train our PCL-Former. Further, we ablated our method in detail to assess the impact of different tIoU thresholds in NMS, the impact of individual modules, and the impact of segment length. It is worth mentioning here that the Temporal Intersection Over Union (tIoU) threshold for NMS is set to 0.4 for THUMOS14 and 0.5 for ActivityNet1.3 and the HACS dataset.
\subsection{Comparison with State-of-the-Art}
\label{subsec:SOTA_Comparison}
\indent \indent  \textbf{THUMOS14:} The quantitative results tabulated in Table~\ref{tab:result_thumosandanet} presents a detailed comparative analysis of our method against several state-of-the-art approaches over the THUMOS14 dataset, evaluating mAP values across varying IoU thresholds from 0.1 to 0.7, i.e., \{0.1:0.1:0.7\}. Notably, our method outperforms the state-of-the-art TAL methods by obtaining the best mAP values with I3D \cite{carreira2017quo} backbone and the second-best mAP values with TSN \cite{wang2016temporal} backbone.\\ \\
\indent \textbf{ActivityNet-1.3:} The quantitative results listed in the rightmost column of Table~\ref{tab:result_thumosandanet} also compare our method (with I3D backbone) against the state-of-the-art TAL approaches on ActivityNet-1.3 dataset. The mAP values are compared using three different tIoU thresholds, including 0.5, 0.75, and 0.95. Our method outperforms the state-of-the-art TAL methods, specifically recent works that include TALLFormer \cite{cheng2022tallformer}, ActionFormer \cite{zhang2022actionformer}, ASL \cite{shao2023action}, TriDet \cite{shi2023tridet}, Re$^2$TAL \cite{zhao2023re2tal}, and ViT-TAD \cite{yang2023adapting} method, by obtaining the best average mAP of 38.2.\\ \\
\indent \textbf{HACS:} Table~\ref{tab:result_hacs} presents the comparison of our method with the state-of-the-art TAL methods on HACS dataset. The mAP values are compared using three different tIoU thresholds that include 0.5, 0.75, and 0.95. The tabulated mAP values demonstrate the effectiveness of our method, outperforming the state-of-the-art TAL methods on the HACS dataset by obtaining the best mAP values across all tIoU thresholds. Overall, our proposed method outperforms the existing TAL methods on THUMOS14, ActivityNet-1.3, and the HACS dataset. These improvements demonstrate the efficacy and potential of our approach for localizing/detecting actions in long-untrimmed videos.
\begin{table}[t]
 \centering
 \caption{\label{tab:result_thumosandanet} \textbf{Comparison} of our \textbf{PCL-Former} with state-of-the-art TAL methods on \textbf{THUMOS14} and \textbf{ActivityNet-1.3} dataset. Results are compared using mAP (\%) at different tIoU thresholds and average mAP (\%).  Best mAP values are listed in \textbf{bold} and runner-up in \underline{underlined}. }
 \resizebox{1.0\textwidth}{!}
 {
  \scriptsize
  \setlength{\tabcolsep}{2.5pt}
  \begin{tabular}{l|c|c|cccccccc|cccc} \hline
  \multirow{2}{*}{Model} & \multirow{2}{*}{Backbone} & \multirow{2}{*}{Venue} & \multicolumn{8}{c|}{\textbf{THUMOS14}} & \multicolumn{4}{c}{\textbf{ActivityNet1.3}}\tabularnewline
 \cline{4-15}
 & & & 0.1 & 0.2 & 0.3 & 0.4 & 0.5 & 0.6  & 0.7 & Avg. & 0.5 & 0.75 & 0.95 & Avg.\\
   \hline
   S-CNN \cite{shou2016temporal} & --- & CVPR &47.7 & 43.5 &36.3 &28.7 &19.0 &--- &--- &19.0 &--- &--- &--- &--- \tabularnewline \cline{4-15}
   SSN \cite{zhao2017temporal} & TSN & ICCV &60.3 & 56.2 &50.6 &40.8 &29.1 &--- &--- &47.4 &43.2 &28.7 &5.6 &25.8 \tabularnewline \cline{4-15}
   TAG \cite{xiong2017pursuit} & TSN & ARXIV &64.1 & 57.7 &48.7 &39.8 &28.2 &--- &--- &47.7 &41.1 &24.1 &5.0 &23.4 \tabularnewline \cline{4-15}
   TALNet \cite{chao2018rethinking} & I3D & CVPR &59.8 & 57.1 &53.2 &48.5 &42.8 &33.8 &20.8 &45.1 &38.2 &18.3 &1.3 &19.2 \tabularnewline \cline{4-15}
   BSN \cite{lin2018bsn} & TSN & ECCV &--- &--- &53.5 & 45.0 &36.9 &28.4 &20.0 &36.8 &46.5 &30.0 &8.0 &28.1 \tabularnewline \cline{4-15}
   GTAN \cite{long2019gaussian} & P3D & CVPR &69.1 & 63.7 &57.8 &47.2 &38.8 &--- &--- &55.3 &52.6 &34.1 &8.9 &31.8 \tabularnewline \cline{4-15}
   BMN \cite{lin2019bmn} & TSN & ICCV &56.0 & 47.4 &38.8 &29.7 &20.5 &--- &--- &38.5 &50.1 &34.8 &8.9 &31.2 \tabularnewline \cline{4-15}
   PGCN \cite{zeng2019graph} & I3D & ICCV &--- & --- &--- &--- &--- &--- &--- &48.3 &33.2 &31.1 &3.3 &22.5 \tabularnewline \cline{4-15}
   GTAD \cite{xu2020g} & TSN & CVPR &66.4 & 60.4 &51.6 &37.6 &22.9 &--- &--- &47.8 &50.4 &34.6 &9.0 &31.3 \tabularnewline \cline{4-15}
   BC-GNN \cite{bai2020boundary} & I3D & ECCV &--- & --- &--- &--- &--- &--- &--- &---  &50.6 &34.8 &9.4 &31.6 \tabularnewline \cline{4-15}
   BU-TAL \cite{zhao2020bottom} & TSN & ECCV &53.9 & 50.7 &45.4 &38.0 &28.5 &--- &--- &43.3 &43.5 &33.9 &9.2 &28.8 \tabularnewline \cline{4-15}
   A2Net \cite{yang2020revisiting} & I3D & TIP &61.1 & 60.2 &58.6 &54.1 &45.5 &32.5 &17.2 &47.0 &43.6 &28.7 &3.7 &25.3 \tabularnewline \cline{4-15}
   PCG-TAL \cite{su2020pcg} & I3D & TIP &71.2 & 68.9 &65.1 &59.5 &51.2 &--- &--- &63.2 &44.3 &29.9 &5.5 &26.5 \tabularnewline \cline{4-15}
   BSN++ \cite{su2021bsn++} & TSN & AAAI &--- &--- &59.9 & 49.5 &41.3 &31.9 &22.8 &41.1 & 51.3&35.7 &9.0 &32.0 \tabularnewline \cline{4-15}
   TCA-Net \cite{qing2021temporal} & TSN & CVPR &--- &--- &60.6 & 53.2 &44.6 &36.8 &26.7 &44.4 &52.3 &36.7 &6.9 &31.9 \tabularnewline \cline{4-15}
   RTD-Action \cite{tan2021relaxed} & TSN & ICCV &--- &--- &68.3 & 62.3 &51.9 &38.8 &23.7 &49.0 &47.2 &30.7 &8.6 &28.8 \tabularnewline \cline{4-15}
   ContexLoc \cite{zhu2021enriching} & I3D & ICCV &--- &--- &68.3 & 63.8 &54.3 &41.8 &26.2 &50.9 &56.0 &35.2 &3.6 &34.2 \tabularnewline \cline{4-15}
   VSGN \cite{zhao2021video} & TSN & ICCV &--- &--- &66.7 & 60.4 &52.4 &41.0 &30.4 &50.2 & 52.3& 35.2 &8.3 & 31.9 \tabularnewline \cline{4-15}
   AFSD \cite{lin2021learning} & I3D & CVPR &--- &--- &67.3 & 62.4 &55.5 &43.7 &31.1 &52.0 &52.4 & 35.3 & 6.5 & 31.4 \tabularnewline \cline{4-15}
   Sub-Action \cite{wang2021exploring} & I3D & TCSVT &66.1 & 60.0 &52.3 &43.2 &32.9 &--- &--- &50.9 &37.1 &24.1 &5.8 &22.3 \tabularnewline \cline{4-15}
   RCL \cite{wang2022rcl} & TSN & CVPR &--- &--- &70.1 & 62.3 &52.9 &42.7 &30.7 &51.7 &--- &--- &--- &--- \tabularnewline \cline{4-15}
   DCAN \cite{chen2022dcan} & TSN & AAAI &--- &--- &68.2 & 62.7 &54.1 &43.9 &32.6 &52.3 &--- &--- &--- & ---\tabularnewline \cline{4-15}
   SRF-Net \cite{ou2022srfnet} & I3D & JSC &--- &53.9 &56.5 & 50.7 &44.8 &33.0 &20.9 &41.2 &--- &--- &--- &--- \tabularnewline \cline{4-15}
   TAL-MTS \cite{gao2022temporal} & I3D & ARXIV &75.3 &73.8 &70.5 &65.0 &56.9 &46.0 &32.7 &61.2 & 52.4 & 34.7 & 6.0 &31.0 \tabularnewline \cline{4-15}
   TadTR \cite{liu2022end} & I3D & TIP &--- &--- &74.8 &69.1 &60.1 &46.6 &32.8 &56.7 & 53.6 & 37.5 & \underline{10.5} &36.7 \tabularnewline \cline{4-15}
   TALLFormer \cite{cheng2022tallformer} & I3D & ECCV &--- &--- &68.4 & --- &57.6 &--- &30.8 &53.9 & 41.3 & 27.3 & 6.3 &24.9 \tabularnewline \cline{4-15}
   ActionFormer \cite{zhang2022actionformer} & I3D & ECCV &--- &--- &82.1 & 77.8 &71.0 &59.4 &43.9 &66.8 & 53.5 & 36.2 & 8.2 &35.6 \tabularnewline \cline{4-15} 
   TriDet \cite{shi2023tridet} & I3D$\mid$R(2+1)D & CVPR & --- & ---&83.6 & 80.1 & 72.9 & 62.4 & 47.4 & 69.3 & 54.7 & 38.0 & 8.4 & 36.8 \tabularnewline \cline{4-15}
    ASL \cite{shao2023action} &I3D & ICCV & --- & --- & 83.1 & 79.0 & 71.7 & 59.7 & 45.8 & 67.9 & 54.1 & 37.4 & 8.0 & 36.2 \tabularnewline \cline{4-15} 
    Re$^{2}$TAL \cite{zhao2023re2tal} & I3D & CVPR & --- & --- &77.4 & 72.6 &64.9 &53.7 &39.0 &--- & 55.2 & 37.8 & 9.0 &37.0 \tabularnewline \cline{4-15}
    ViT-TAD \cite{yang2023adapting} & ViT-B & ARXIV &--- &---&85.1 & 80.9 &74.2 &61.8 &45.4 &69.5 & 55.8 &38.4 & 8.8 &37.4 \tabularnewline \cline{4-15}
    LFA \cite{tang2024learnable} & I3D & TIP &--- &---&83.0 & 79.5 & 73.8 & 62.5 &  48.2 & 69.4 & 54.7 & 37.3 & 8.6 & 36.6 
    \tabularnewline \cline{4-15}
    EEI \cite{mokari2024enhancing} & TSN & IVC & 66.2 & 62.6& 56.7 & 48.8 & 39.3 & --- &  --- & --- & 37.8 & 28.3 & 5.1 & 27.2
    \tabularnewline \cline{4-15}
    DeTAL \cite{li2024detal} & I3D & TPAMI & --- & --- & 39.8 & 33.6 & 25.9 & 17.4 &  9.9 & 25.3 & 39.3 & 26.4 & 5.0 & 25.8
    \tabularnewline \cline{4-15}
    EPN \cite{wu2024ensemble} & I3D & TNNLS & 72.5 & 67.1 & 58.5 & 48.0 & 37.1 & 25.7 &  13.6 & --- & 41.4 & 25.1 & 6.2 & 25.2
    \tabularnewline \cline{4-15}
    NGPLG \cite{li2024neighbor} & I3D & TIP & 77.9 & 73.9 & 66.6 & 59.4 & 48.6 & 36.7 &  22.7 & 55.1 & 41.3 & 30.9 & 4.8 & 26.5
    \tabularnewline \cline{4-15}
    ISSF \cite{yun2024weakly} & I3D & AAAI & 72.4 & 66.9 & 58.4 & 49.7 & 41.8 & 25.5 &  12.8 & 46.8 & 39.4 & 25.8 & 6.4 & 25.8
    \tabularnewline \cline{4-15}
    CausalTAD \cite{liu2024harnessing} & I3D$\mid$TSP & ARXIV & --- & --- & 84.4 & 80.7 & 73.5 & 62.7 & 47.4 & 69.7 & 55.6 & 38.5 & 9.4 & 37.4
     \tabularnewline \cline{4-15}
    AdaTAD \cite{liu2024end} & SlowFast-R50 & CVPR & --- & --- & 84.5 & 80.2 & 71.6 & 60.9 &  46.9 & 68.8 & 56.1 & 38.9 & 9.0 & 37.8
    \tabularnewline \cline{1-15}
   PCL-Former (\textbf{ours})  & TSN & ---&\underline{87.9} &\underline{86.4} &\underline{85.7} &\underline{81.4} &\underline{75.3} &\underline{63.9} &\underline{49.1} &\underline{75.6} & 55.7 & 38.4 & 9.7 & 37.6 \tabularnewline \cline{4-15}
   PCL-Former (\textbf{ours})  & I3D & ---&\textbf{89.6} &\textbf{88.3} &\textbf{86.8} &\textbf{84.5} &\textbf{77.2} &\textbf{68.6} &\textbf{54.1} &\textbf{78.4} &\textbf{56.3} &  \textbf{39.1} & \textbf{11.7} &\textbf{38.6} \tabularnewline \hline
  \end{tabular}
 }
\end{table} 
\subsection{Ablation Study}  
\begin{wraptable}{r}{0.72\textwidth}
  \centering
  \caption{ \label{tab:result_hacs} \textbf{Comparison} of our PCL-Former (I3D backbone) with state-of-the-art TAL methods on \textbf{HACS} dataset. The results are compared using mAP(\%) at different tIoU thresholds and average mAP(\%).}
  \label{tab:example}
 \begin{tabular}{@{}l|c|c|c|c@{}} \hline
    Method &  0.5 & 0.75 & 0.95 & Avg \tabularnewline
    \cline{1-5}
S-2D-TAN \cite{zhang2019learning} &  - & - & - & 23.4 \tabularnewline
GTAD \cite{xu2020g} &  - & - & -  & 27.5\\
BMN \cite{lin2019bmn} &  52.5 & 36.4 & 10.4  & 35.8\\
TALL-Former \cite{cheng2022tallformer} &  55.0  & 36.1 & 11.8  & 36.5\\
TadTR \cite{liu2022end} &  47.1  & 32.1 & 10.9  & 30.0\\
TCA-NET \cite{qing2021temporal} &  \underline{56.7} & \underline{41.1} & \underline{12.1}  & \underline{39.7}\\
TriDet-I3D \cite{shi2023tridet} &  54.5 & 36.8 & 11.5  & 36.8\\
TriDet-SlowFast \cite{shi2023tridet} &  56.7 & 39.3 & 11.7  & 38.6\\
\hline
PCL-Former (\textbf{ours}) & \textbf{61.2} & \textbf{47.9 } & \textbf{16.4}  & \textbf{44.5}\\
\hline
  \end{tabular}
\end{wraptable} 
\textbf{Impact of Different tIoU Thresholds in NMS:} To 
evaluate the impact of tIoU threshold on overall performance of our method, we follow the settings in \cite{xu2017r} and \cite{gao2022temporal} and examine the performance of our method with four different tIoU thresholds that include \{0.2, 0.3, 0.4, and 0.5\} in NMS on THUMOS14, ActivityNet-1.3, and HACS dataset. Table~\ref{tab:tiou_thumos}, 
\ref{tab:tiou_activitynet}, and \ref{tab:tiou_hacs} illustrate the impact of various tIoUs in NMS on the overall performance of PCL-Former on THUMOS14, ActivityNet-1.3, and HACS dataset, respectively. \\ \\
\indent \textbf{Impact of Individual Modules:} To study the effect of individual modules, we investigated the performance of PCL-Former on THUMOS14, ActivityNet-1.3, and HACS dataset under four different settings including CL-Former (excluding Proposal-Former), PC-Former (excluding Localization-Former), PCL-Former$_{(w/o BRM)}$, and PCL-Former$_{(complete)}$. The obtained quantitative results in terms of average mAP values are presented in Table~\ref{tab-indi_mod}. The first setting (i.e., CL-Former) excludes Proposal-Former and operates directly on a sliding window. The second setting (i.e., PC-Former) excludes Localization-Former and operates on foreground sequence (prediction from Proposal-Former) to predict the action. The action class probabilities are utilized to predict the action boundaries. Third setting PCL-Former$_{(w/o BRM)}$ does not utilize the BRM module in a post-processing step. The last setting utilizes the complete PCL-Former architecture for temporal action localization tasks. The listed mAP values in Table~\ref{tab-indi_mod} demonstrate the effectiveness of the individual module in PCL-Former architecture.\\ \\ 
\begin{table}[t]\centering
\caption{\textbf{Impact} of different tIoU thresholds in NMS on the performance of our \textbf{PCL-Former}, over (a) THUMOS14, (b) ActivitNet-1.3, and (c) HACS dataset. \label{tab-tiou_impact}} 
\subfloat[\textbf{Impact} of different tIoU thresholds in NMS on mAP values (\%) over \textbf{THUMOS14} dataset. \label{tab:tiou_thumos}]
{
  \scriptsize
  \setlength{\tabcolsep}{8.5pt}
  \begin{tabular}{c|c|c|c|c|c|c|c|c|c} \hline
    \multirow{2}{*}{tIoU Thresholds} & \multicolumn{9}{c}{mAP (\%)} \\ 
    \cline{2-10}
    & &0.1&0.2&0.3&0.4&0.5&0.6&0.7&Avg\\
    \hline
     0.2& \multirow{5}{*}{\rotatebox[origin=c]{90}{TSN}} &86.2&85.4&84.6&80.5&74.3&62.5&47.7&74.4\\ 
    0.3&&87.4&\underline{86.1}&\underline{85.5}&80.7&74.4&63.4&48.5&75.1\\
    0.4& & \textbf{87.9}&\textbf{86.4}&\textbf{85.7}&\textbf{81.4}&\textbf{75.3}&\textbf{63.9}&\textbf{49.1}&\textbf{75.6}\\
    0.5&&\underline{87.8}&85.9&85.4&\underline{80.8}&\underline{74.5}&\underline{63.5}&\underline{48.9}&\underline{75.2} \\ 
    \hline
    0.2& \multirow{5}{*}{\rotatebox[origin=c]{90}{I3D}} &88.3&87.7&86.1&83.4&76.4&67.5&52.6&77.4\\ 
    0.3&&88.8&88.0&\underline{86.7}&84.1&76.7&68.1&\underline{53.5}&77.9\\
    0.4&&\textbf{89.6}&\textbf{88.3}&\textbf{86.8}&\textbf{84.5}&\textbf{77.2}&\textbf{68.6}&\textbf{54.1}&\textbf{78.4}\\
    0.5&&\underline{89.0}&\underline{88.2}&86.4&\underline{84.2}&\underline{76.8}&\underline{68.4}&53.4&\underline{78.1} \\ \hline
\end{tabular}
}
\\
\subfloat[\textbf{Impact} of different tIoU thresholds in NMS on mAP values (\%) over \textbf{ActivityNet-1.3} dataset. \label{tab:tiou_activitynet}]
{
\resizebox{0.40\textwidth}{!}{
    \begin{tabular}{c|c|c|c|c} \hline
    \multirow{2}{*}{tIoU Thresholds} & \multicolumn{4}{c}{mAP (\%)} \\ 
    \cline{2-5}
    &0.5&0.75&0.95&Avg\\
    \hline
   0.2&\underline{55.7}&37.8&10.1&37.0\\ 
    0.3&55.2&\underline{38.2}&\underline{10.8}&\underline{37.2}\\ 
    0.4&55.5&38.1&10.4&37.1\\
    0.5&\textbf{56.3}&\textbf{39.1}&\textbf{11.7}&\textbf{38.6}\\ \hline
\end{tabular}
}
}~
\subfloat[\textbf{Impact} of different tIoU thresholds in NMS on mAP values (\%) over \textbf{HACS} dataset. \label{tab:tiou_hacs}]
{
\resizebox{0.40\textwidth}{!}{
    \begin{tabular}{c|c|c|c|c} \hline
    \multirow{2}{*}{tIoU Thresholds} & \multicolumn{4}{c}{mAP (\%)} \\ 
    \cline{2-5}
    &0.5&0.75&0.95&Avg\\
    \hline
    0.2&60.3&46.8&15.8&43.6\\ 
    0.3&\underline{60.9}&47.1&\underline{16.1}&\underline{44.0}\\ 
    0.4&60.6&\underline{47.3}&16.0&43.9\\ 0.5&\textbf{61.2}&\textbf{47.9}&\textbf{16.4}&\textbf{44.5}\\ \hline
\end{tabular}
}
}
\end{table}
\begin{table}[t]
\caption{\textbf{Impact} of individual modules of PCL-Former on the overall performance in terms of Avg mAP on \textbf{THUMOS14}, \textbf{ActivityNet-1.3}, and \textbf{HACS}  dataset. Here, P-Former, C-Former, and L-Former represent Proposal-Former, Classification-Former, and Localization-Former, respectively. \label{tab-indi_mod}} 
 \centering
  \scriptsize
  \setlength{\tabcolsep}{8.5pt}
  \begin{tabular}{l|c|cccc|c} \hline
  Method             & Backbone &P-Former & C-Former & L-Former & BRM &Avg  \\
  \hline
  \multicolumn{7}{c}{\textbf{THUMOS14}} \\ \hline
  CL-Former  & \multirow{5}{*}{\rotatebox[origin=c]{90}{TSN}}    &   & \checkmark  &\checkmark   & \checkmark     & 72.0 \\
  PC-Former      &  & \checkmark  &  \checkmark    &  & \checkmark     & 70.2 \\
  PCL-Former$_{(w/o BRM)}$     &  & \checkmark  & \checkmark  & \checkmark  &   & \underline{73.4}\\
  PCL-Former$_{(complete)}$     &  & \checkmark  & \checkmark  & \checkmark  & \checkmark   & \textbf{75.6}\\\hline
  CL-Former  & \multirow{5}{*}{\rotatebox[origin=c]{90}{I3D}}    &   & \checkmark  &\checkmark   & \checkmark     & 75.3 \\
  PC-Former      &  & \checkmark  &  \checkmark    &  & \checkmark     & 73.2 \\
  PCL-Former$_{(w/o BRM)}$     &  & \checkmark  & \checkmark  & \checkmark  &   & \underline{76.8}\\
 PCL-Former$_{(complete)}$     &  & \checkmark  & \checkmark  & \checkmark  & \checkmark   & \textbf{78.4}\\ \hline
  \multicolumn{7}{c}{\textbf{ActivityNet-1.3}} \\ \hline
  CL-Former  & \multirow{5}{*}{\rotatebox[origin=c]{90}{I3D}}    &   & \checkmark  &\checkmark   & \checkmark     & 35.1 \\
  PC-Former      &  & \checkmark  &  \checkmark    &  & \checkmark     & 33.0 \\
  PCL-Former$_{(w/o BRM)}$     &  & \checkmark  & \checkmark  & \checkmark  &   & \underline{36.6}\\
  PCL-Former$_{(complete)}$     &  & \checkmark  & \checkmark  & \checkmark  & \checkmark   & \textbf{38.6}\\ \hline
  \multicolumn{7}{c}{\textbf{HACS}} \\ \hline
  CL-Former  & \multirow{5}{*}{\rotatebox[origin=c]{90}{I3D}}    &   & \checkmark  &\checkmark   & \checkmark     & 41.4 \\
  PC-Former      &  & \checkmark  &  \checkmark    &  & \checkmark     & 39.3 \\
  PCL-Former$_{(w/o BRM)}$     &  & \checkmark  & \checkmark  & \checkmark  &   & \underline{42.9}\\
  PCL-Former$_{(complete)}$     &  & \checkmark  & \checkmark  & \checkmark  & \checkmark   & \textbf{44.5}\\ \hline
  \end{tabular}
\end{table}
\indent \textbf{Impact of Segment Length:} To investigate the impact of segment length on the performance of our proposed PCL-Former, we conducted experiments with four different segment lengths including 16, 32, 64, and 128 over THUMOS14, ActivityNet-1.3, and HACS dataset. The obtained results with different segment lengths are presented in Table~\ref{tab:segment_thumos}, \ref{tab:segment_activitynet}, and \ref{tab:segment_hacs}. Observing the results on the THUMOS14 dataset in Table~\ref{tab:segment_thumos}, it can be noticed that our PCL-Former (with both backbones TSN and I3D) obtains the best average mAP values using segment length of 64 and second-best average mAP values with segment length 32. Similarly, on ActivityNet-1.3 and HACS dataset in Table~\ref{tab:segment_activitynet} and  \ref{tab:segment_hacs}, the proposed method achieves optimal results using a segment length of 64 (annotated with $\ast$ in Table \ref{tab:segment_thumos}, \ref{tab:segment_activitynet}, and \ref{tab:segment_hacs}) and obtains the second-best average mAP values with a segment length of 32.
\begin{table}[t]\centering
\caption{\textbf{Impact} of different temporal lengths of segments on the performance of our \textbf{PCL-Former}, over (a) THUMOS14, (b) ActivitNet-1.3, and (c) HACS dataset. \label{tab-tiou_impact}} 
\subfloat[\textbf{Impact} of different segment lengths on mAP values (\%) over \textbf{THUMOS14} dataset. \label{tab:segment_thumos}]
{
  \scriptsize
  \setlength{\tabcolsep}{8.5pt}
  \begin{tabular}{c|c|c|c|c|c|c|c|c|c} \hline
    \multirow{2}{*}{Segment Length} & \multicolumn{9}{c}{mAP (\%)} \\ 
    \cline{2-10}
    & &0.1&0.2&0.3&0.4&0.5&0.6&0.7&Avg\\
    \hline
    16& \multirow{5}{*}{\rotatebox[origin=c]{90}{TSN}} & 86.2& 83.8& 83.3& 78.8& 72.2& 63.0 & 47.1 & 73.4\\ 
    32& & \underline{87.6} & \underline{85.1} & \underline{84.9} & \underline{79.7}& \underline{73.9} & \underline{63.6} & \underline{47.8} & \underline{74.6}\\
     64($\ast$)&  &\textbf{87.9} & \textbf{86.4} & \textbf{85.7} & \textbf{81.4} & \textbf{75.3} & \textbf{63.9} & \textbf{49.1} & \textbf{75.6} \\
    128& & 86.7& 84.6& 83.6 & 79.4& 73.4& 63.5 & 47.7& 74.1\\ 
    \hline
    16& \multirow{5}{*}{\rotatebox[origin=c]{90}{I3D}} & 86.9 & 86.2 & 84.6 & 81.4 & 74.9 & 66.3 & 51.0 & 75.9\\ 
    32& & \underline{88.3}& \underline{87.1} & \underline{85.2} & \underline{83.3} & \underline{76.1} & \underline{67.4} & \underline{51.8} & \underline{77.0}\\
     64($\ast$)& &\textbf{89.6} & \textbf{88.3} & \textbf{86.8} & \textbf{84.5} & \textbf{77.2} & \textbf{68.6} & \textbf{54.1} & \textbf{78.4} \\
    128& & 87.3& 86.0 & 85.0 & 81.8 & 76.0 & 67.0 & 51.4& 76.3\\ \hline
\end{tabular}
}
\\
\subfloat[\textbf{Impact} of different segment lengths on mAP values (\%) over \textbf{ActivityNet-1.3} dataset. \label{tab:segment_activitynet}]
{
  \resizebox{0.40\textwidth}{!}{
    \begin{tabular}{c|c|c|c|c} \hline
    \multirow{2}{*}{Segment Length} & \multicolumn{4}{c}{mAP (\%)} \\ 
    \cline{2-5}
    &0.5&0.75&0.95&Avg\\
    \hline
    16& 53.4 &35.6 &10.2 & 35.5 \\ 
    32& \underline{55.5} & \underline{38.3} & \underline{11.3} & \underline{37.5} \\ 
    64($\ast$)& \textbf{56.3} & \textbf{39.1} & \textbf{11.7} & \textbf{38.6} \\ 
    128&  53.8 &37.2 & 10.9 & 36.4\\  \hline
    \end{tabular}
  }
}~
\subfloat[\textbf{Impact} of different segment lengths on mAP values (\%) over \textbf{HACS} dataset. \phantom{This text is invisible} \label{tab:segment_hacs}]
{
  \resizebox{0.40\textwidth}{!}{
    \begin{tabular}{c|c|c|c|c} \hline
    \multirow{2}{*}{Segment Length} & \multicolumn{4}{c}{mAP (\%)} \\ 
    \cline{2-5}
    &0.5&0.75&0.95&Avg\\
    \hline
    16& 58.6& 46.3 & 14.3 & 42.4\\ 
    32& \underline{59.9} & \underline{47.5} & \underline{14.8} & \underline{43.4}\\ 
    64($\ast$)& \textbf{61.2} & \textbf{47.9} &\textbf{16.4} & \textbf{44.5} \\ 
    128& 59.2 & 46.1& 14.6 & 42.6\\ \hline
    \end{tabular}
  }
}
\end{table}
\section{Conclusion}
\label{sec:conclusion}
\indent \indent Precise temporal action localization in untrimmed videos remains a critical yet challenging task in the context of human action detection. Untrimmed videos typically contain multiple action instances, often interspersed with irrelevant content such as background scenes or unrelated activities. To address these complexities, we thoroughly explore the inherent rich spatio-temporal properties of transformers in multi-stage settings for temporal action localization. This exploration leads to the development of a novel, three-tier transformer-based architecture, named PCL-Former, designed specifically for the temporal action localization task. PCL-Former consists of three task-specific transformer modules, each paired with a dedicated loss function to handle distinct subtasks. Specifically, the Proposal-Former identifies candidate segments that may contain actions, the Classification-Former classifies the action category in a given segment, and the Localization-Former predicts the temporal boundaries of the action instance (start and end of action). During training, each sub-network is independently trained with its respective loss, while inferencing, the Classification-Former, and Localization-Former operate on the candidate segments identified by the Proposal-Former. Extensive experiments on THUMOS14, ActivityNet-1.3, and HACS demonstrate that PCL-Former outperforms state-of-the-art TAL methods on all datasets. Additionally, we conduct thorough ablation studies, analyzing the impact of different IoU thresholds in NMS, the impact of individual module, and the impact of segment length. The resulting mAP values provide deeper insights into the architecture’s configuration and the effectiveness of its components.

\section*{Acknowledgment} 
This research was supported in part by the Air Force Office of Scientific Research (AFOSR) Contract Number FA9550-22-1-0040. The authors would like to acknowledge Dr. Erik Blasch from the Air Force Research Laboratory (AFRL) for his guidance and support on the project. Any opinions, findings, and conclusions or recommendations expressed in this material are those of the author(s) and do not necessarily reflect the views of the Air Force, the Air Force Research Laboratory (AFRL), and/or AFOSR.

\newpage

\bibliography{iclr2025_conference}
\bibliographystyle{iclr2025_conference}

\appendix

\section{Appendix to Experiments}
\subsection{Classwise Accuracy Comparison with State-of-the-art}
\begin{figure}[h]
    \centering
    \includegraphics[width=1.0\linewidth]{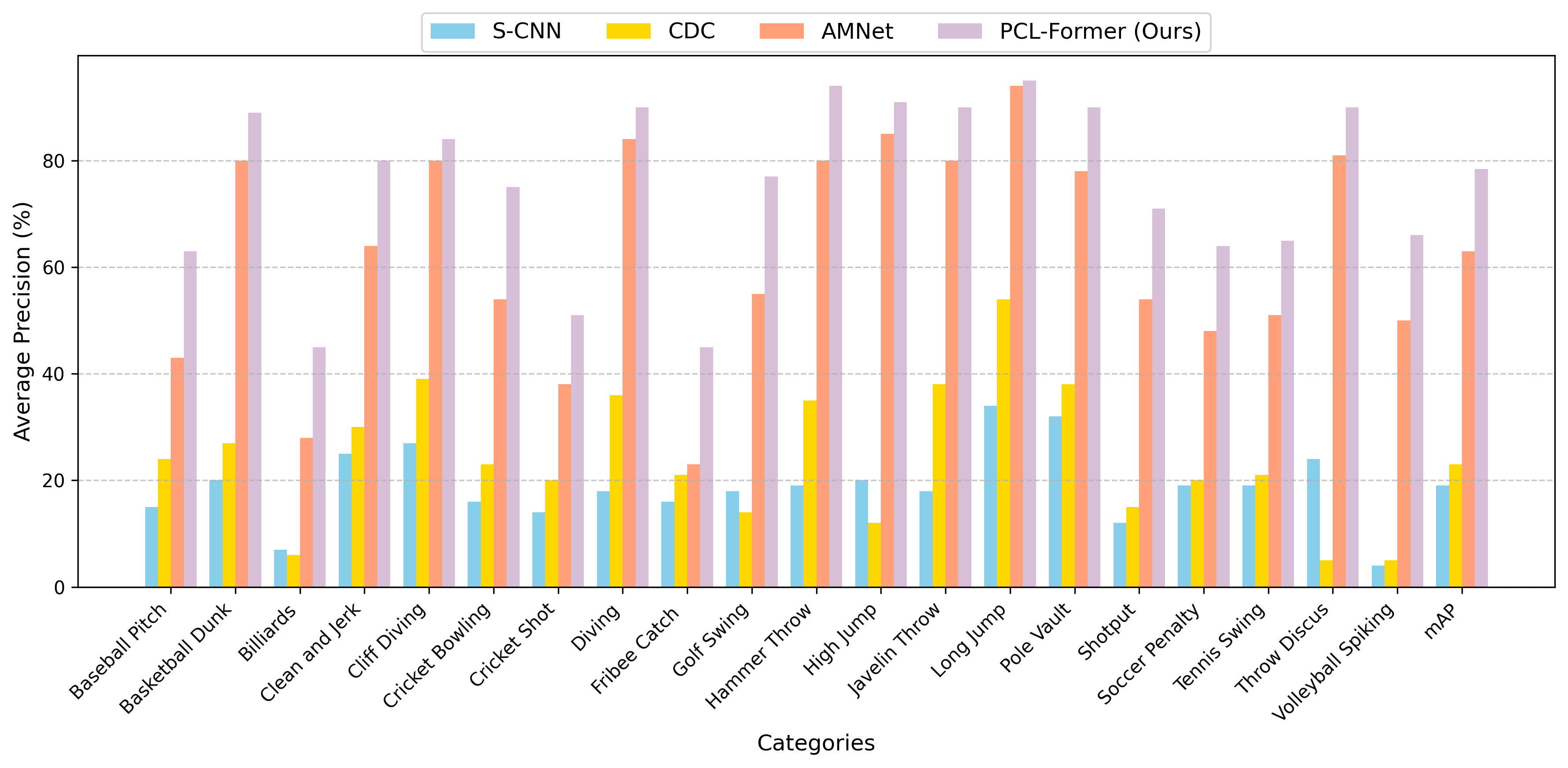}
    \caption{Class-wise performance comparison with S-CNN \cite{shou2016temporal}, CDC \cite{shou2017cdc}, and AMNet \cite{kang2023action} methods on THUMOS14 dataset. The results are evaluated using AP@Avg across various tIoU thresholds.}
    \label{fig:class-wise_acc}
\end{figure}

In addition to the quantitative comparison given in Section 4.4 (Comparison with State-of-the-Art), we further evaluated our PCL-Former by comparing its class-wise performance with the S-CNN \cite{shou2016temporal}, CDC \cite{shou2017cdc}, and AMNet \cite{kang2023action} on THUMOS14 dataset, as illustrated in Figure~\ref{fig:class-wise_acc}. As shown in Figure~\ref{fig:class-wise_acc}, our PCL-Former outperforms all other TAL methods by a significant margin, achieving the highest Average Precision (AP) across all classes. In terms of Mean Average Precision (mAP), our method achieves the highest performance, with a mAP of 78.4\% followed by AMNet \cite{kang2023action}, which attains the second-best mAP of 63.3\%, demonstrating a substantial gap between our approach and the competing methods. It is worth mentioning here that all the references cited in this supplementary file are directly sourced from the main paper. This ensures consistency and allows readers to easily trace the original studies, methodologies, and related work discussed throughout the supplementary material.

\subsection{Qualitative Results}
\begin{figure}[t]
    \centering
    \includegraphics[width=1.0\linewidth]{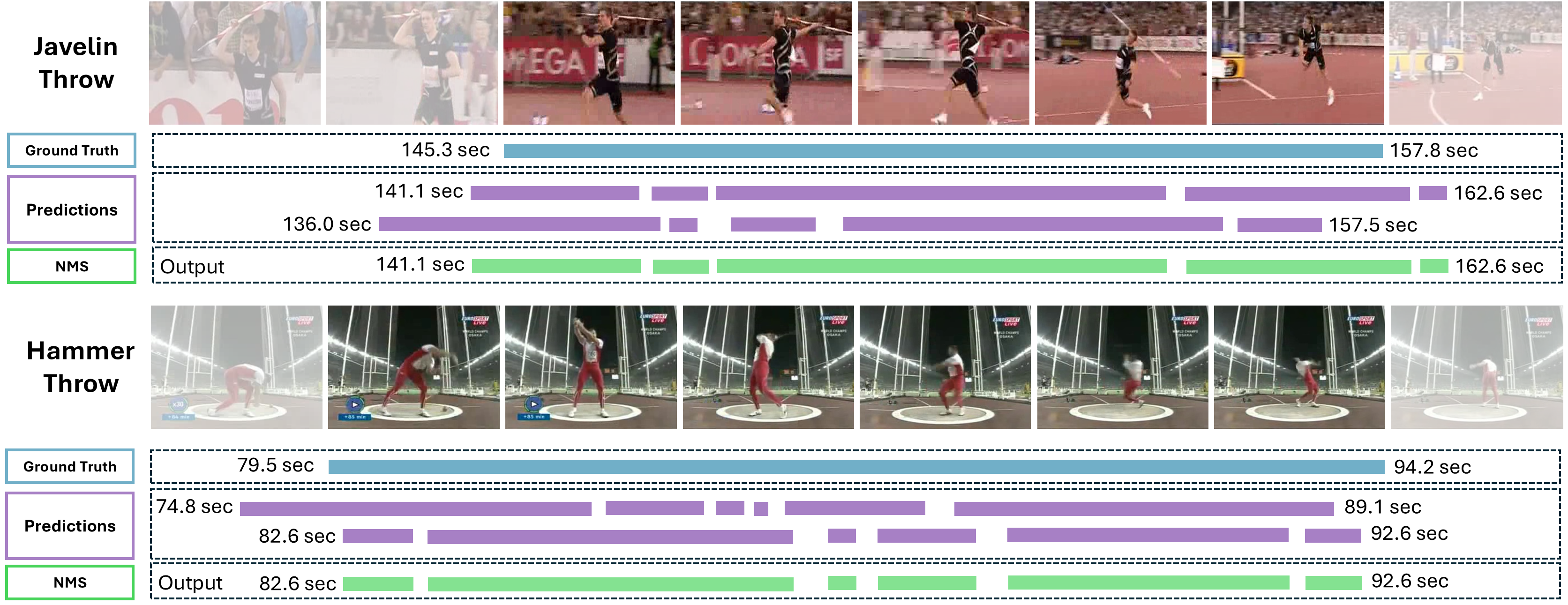}
    \caption{The prediction results for the \textbf{Javelin Throw} and \textbf{Hammer Throw} action instances from the THUMOS14 test set, evaluated using an overlap threshold of 0.4. For each ground truth instance, two predictions are provided. The NMS process is applied to both, selecting the prediction with the highest overlap with the ground truth.}
    \label{fig:loc_vis-1}
\end{figure}

\begin{figure}[t]
    \centering
    \includegraphics[width=1.0\linewidth]{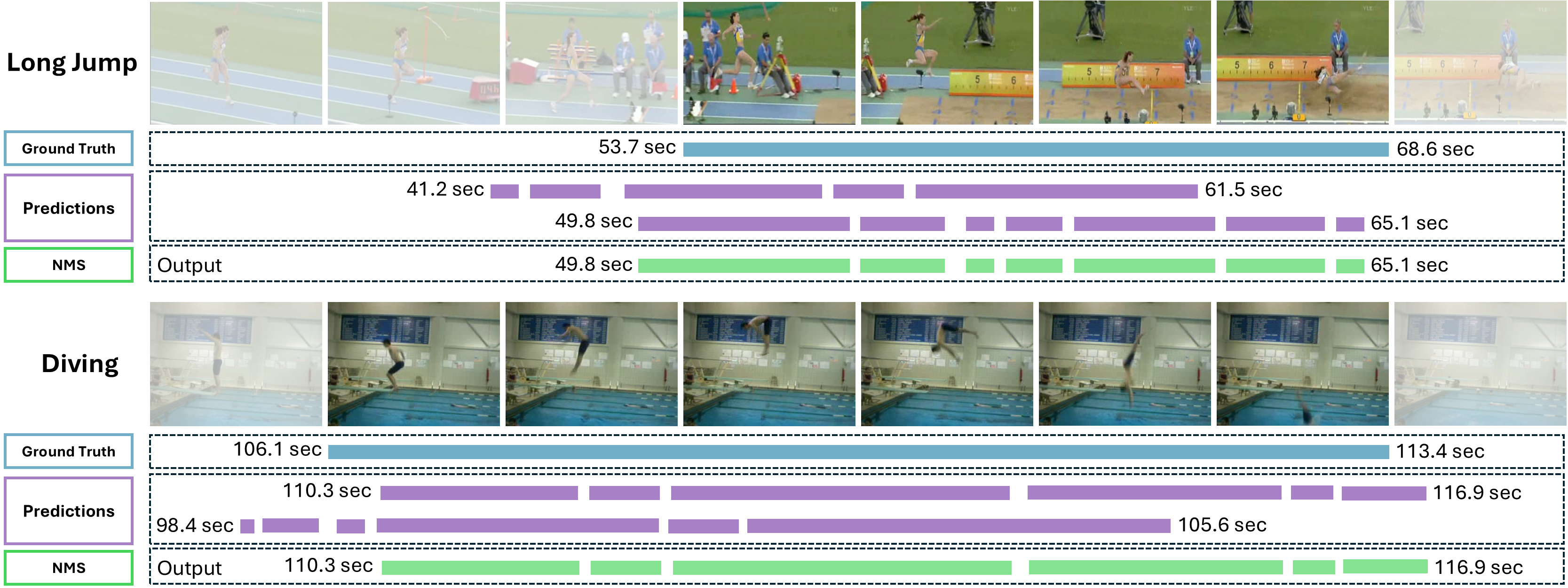}
    \caption{The prediction results for the \textbf{Long Jump} and \textbf{Diving} action instances from the THUMOS14 test set, evaluated using an overlap threshold of 0.4. For each ground truth instance, two predictions are provided. The NMS process is applied to both, selecting the prediction with the highest overlap with the ground truth.}
    \label{fig:loc_vis-2}
\end{figure}

Besides quantitative evaluation, we further assessed our method's performance in terms of temporal boundary prediction on the test set of the THUMOS14 dataset. This qualitative analysis aimed to evaluate the capability of our approach to accurately localize action instances by predicting their start and end times within video sequences. By examining the temporal alignment of predicted segments with ground truth annotations, we highlighted the precision and robustness of our method in handling complex temporal boundaries. For the qualitative evaluation, we selected four videos from the test set of the THUMOS14 dataset, corresponding to the actions \textbf{Javelin Throw}, \textbf{Hammer Throw}, \textbf{Long Jump}, and \textbf{Diving}. An overlap threshold of 0.4 is used during Non-Maximum Suppression (NMS) to identify the optimal temporal predictions for each action instance. The resulting temporal predictions are illustrated in Figure~\ref{fig:loc_vis-1} and Figure~\ref{fig:loc_vis-2}, highlighting the effectiveness of our method in capturing precise action boundaries.
\\ \\
Figure 5 depicts the temporal action localization results for Javelin Throw and Hammer Throw, from the THUMOS14 test set. The visualization includes three components: ground truth annotations, predicted temporal segments, and the final outputs after applying Non-Maximum Suppression (NMS). It is worth mentioning here that, we provided two distinct predictions for each ground truth instance. For example, in the case of the Javelin Throw, the first prediction spans from 141.1 seconds to 162.6 seconds, while the second prediction spans from 136.0 seconds to 157.5 seconds. By applying Non-Maximum Suppression (NMS), the prediction with the highest overlap with the ground truth (spanning from 145.3 seconds to 157.8 seconds) is selected as the optimal prediction, which in this case corresponds to the first prediction. Similarly, for the Hammer Throw, NMS identifies the second prediction as the optimal choice, demonstrating the effectiveness of the selection process in refining temporal action localization results.\\ \\
\indent Further, Figure~\ref{fig:loc_vis-2} illustrates the qualitative results of our method on Long Jump and Diving. Each activity is accompanied by ground truth annotations, two predicted temporal intervals, and the final output selected by Non-Maximum Suppression (NMS). For the Long Jump, the ground truth spans from 53.7 seconds to 68.6 seconds. Among the two predicted intervals, spanning 41.2 seconds to 61.5 seconds and 49.8 seconds to 65.1 seconds, NMS identifies the second prediction as optimal due to its higher overlap with the ground truth. Similarly, in the Diving activity, the ground truth spans from 106.1 seconds to 113.4 seconds. Of the two predicted intervals, spanning 98.4 seconds to 105.6 seconds and 110.3 seconds to 116.9 seconds, NMS selects the second interval as the optimal prediction. These results further emphasize the ability of our approach to refine temporal predictions and accurately align them with the ground truth boundaries.
\end{document}